\documentclass[10pt,twocolumn,letterpaper]{article}

\pdfoutput=1
\usepackage{cvpr}
\usepackage{times}
\usepackage{epsfig}
\usepackage{graphicx}
\usepackage{amsmath}
\usepackage{amssymb}

\usepackage[breaklinks=true,bookmarks=false]{hyperref}
\cvprfinalcopy 

\def\cvprPaperID{****} 

\begin{document}

\title{%
    Pay Attention to Convolution Filters: \\
    \large Towards Fast and Accurate Fine-Grained Transfer Learning \thanks{CS 194 Final Project Report. Team name: ``compressed r-cnn''}
}

\author{Simon Mo \quad\quad Ryan Cheng \quad\quad Ian Fang\\
UC Berkeley\\
{\tt\small \{xmo,chengruizhe,tinyik\} @berkeley.edu}
}

\maketitle

\begin{abstract}
  We propose an efficient transfer learning method for adapting ImageNet pre-trained Convolutional Neural Network (CNN) to fine-grained image classification task. Conventional transfer learning methods typically face the trade-off between training time and accuracy. By adding ``attention module'' to each convolutional filters of the pre-trained network, we are able to rank and adjust the importance of each convolutional signal in an end-to-end pipeline. In this report, we show our method can adapt a pre-trianed ResNet50 for a fine-grained transfer learning task within 10 epochs and achieve accuracy above conventional transfer learning methods and close to models trained from scratch. Our model also offer interpretable result because the rank of the convolutional signal shows which convolution channels are utilized and amplified to achieve better classification result, as well as which signal should be treated as noise for the specific transfer learning task, which could be pruned to lower model size.
\end{abstract}

\section{Introduction}
Our work combines three different topics in deep learning--–transfer learning, fine-grained image classification and attention methods---to solve a practical problem. In this section, we will first introduce the problem we are trying to solve and its implication. We will then introduce the three separate topics that inspired our works. Lastly,  we will briefly summarize our contribution and result.  

\subsection{Problem and Implication}
Most image classification problems in real life are fine-grained classification problems. For example, self-driving cars need classification models that correctly classify different kinds of street signs; medical imaging application requires model to distinguish different types of cells. However, most pre-trained image classification models were trained on ImageNet dataset. The dataset is very coarse grained; e.g. models were tasked to distinguish between trucks and cats. It is widely believed that these models constitute as common feature extractors that can be adapted for specific tasks. 
\emph{
How can we efficiently take advantage of pre-trained network and adapt them for fine-grained image classification problem? 
}

Our objective is two-fold: (1) We want the training process to be as fast as possible. Ideally, a training process should take less than 10 epochs. This can be done via taking advantages of learned weights in the pre-trained network. (2) We want the model accuracy to be as high as possible. It should perform reasonably well compared to a model that's trained from scratch. 

We hope our work can benefit machine learning practitioners in industry. Using our methods, a small network that learned to classify fine-grained objects well can be trained and iterated quickly. For example, an ``expert network'' that is able to distinguish very similar classes can serve as part of a larger ensemble \cite{mixture-of-expert} or inference cascade \cite{idk-cascade} to improve overall accuracy. 

\subsection{Transfer Learning}
High accuracy image classification models like ResNet \cite{ResNet} and Inception-V3 \cite{inception} that are trained on ImageNet \cite{ImageNet_cvpr09} have been made widely available across deep learning frameworks \cite{pytorch-modelzoo}. Transfer learning attempts to use these pre-trained models as starting point to adapt the network for a specific task. A general approach \cite{231n-transfer} is to freeze the weights of some layers, typically low-level features, and re-train the high level features and fully-connected layers. If the new dataset's distribution is similar to that of ImageNet, it is recommended to only re-train the fully connected layer. If the new dataset's distribution is different from ImageNet, it is recommended to only keep the low level features and re-train layers that correspond to high level features.

\subsection{Fine-Grained Image Classification}
Fine-grained image classification tasks are tasks where the objects in the source images are very similar and requires fine-grained features: for example, classifying animal species \cite{inaturalist} requires the network to pick up features like color patterns of frog skins or specific shapes of bird beaks. To solve this problem, the academia has moved away from training model from scratch. Many researchers have started working on adapting learned features efficiently to identity important details. Methods like subset feature learning \cite{fine-grained-1}, mixture of convolutional networks \cite{fine-grained-mixture}, or adding visual attention to image \cite{fine-grained-attention-loc} have been proven to be effective. However, many of these methods are very domain specific, and most importantly, their training still takes a long time.

\subsection{Attention Methods}
Attention is first introduced in the setting of Neural Machine Translation \cite{bahdanau}. Bahdanau et al. applied attention to each state of RNN to jointly produce a weighted vector that captures the latent semantic meaning of the sentence being translated. Xu et al. borrows the same idea and applies it to the task of image captioning \cite{bengio}. By using attention mechanism, they achieved state-of-the-art performance while purposing a novel way of visualizing the regions in the image that are most heavily weighted when the network is generating each word in the caption.

\subsection{Our Contribution}
Our work ties together all three ideas to attack the problem of transfer learning for fine-grained image classification. In particular, we  draw inspiration from attention method in text models and image model to rank convolution filters in a pre-trained network regarding a specific fine-grained transfer learning task. Our goal is to explore a method that strikes a good balance between speed and validation accuracy. 

Our final model follows from the intuition that pre-trained models are over-parameterized. By adding trainable ``attention'' weights to each convolution channel and optimize for the classification loss, these ``attention'' weights will diverge from their initial states (which are just $1$, mapping identical filters). After few iterations, attention weights can either amplify a convolution channel or lower the activation of it. For example, if a convolution channel detecting the vertical stripes pattern from species A contributes greatly to the success of correctly classifying species A, we expect the network would learn to ``pay more attention'' to this channel by amplifying the attention weight corresponds to that channel. 

We also expect the network will learn to ``pay less attention'' to features that are irrelevant/less important to the fine-grained dataset. For example, high level features (like floppy ears of dogs or the shape of a semi-trailer truck) in a ImageNet trained model (like Inception-V3) \cite{distill-viz} are not really useful for a fine-grained dataset.

\subsection{Structure of the Report}

This report will be structured as followed. Section \ref{sec:related-work} will identify several key literature on which we ground our assumptions and methods. Section \ref{sec:setup} will discuss our experimental setup and baseline models. Section \ref{sec:final-model} will discuss detail implementation of our models and training scheme. Section \ref{sec:disc} will identify the set of experiments we ran in order to explore many facets of efficent fine-grained transfer learning. Finally, in section \ref{sec:lesson-learned} and \ref{sec:future-work}, we will summarize our result and explore future works.

\section{Related Work}\label{sec:related-work}
This section is organized by a series of key assumptions we made.

\subsection{Key Assumption 1: Large Networks are Over-Parametrized}
Although almost a consensus among deep learning community, we found interesting theoretical and experimental results from Frankle et al \cite{lottery}. Frankle et al. explore so-called ``The Lottery Ticket Hypothesis" and realized that the success of many large network can be attributed to the large number of layers and parameters that made possible for successful sub-networks to appear, which are usually discovered via pruning. 

\subsection{Key Assumption 2: ImageNet Pre-Trained Convolution Channels are Good Feature Extractors}
In particular, we need to assume that the base network that we adapt for transfer learning purpose is a high quality network in terms of its convolution channels. We assume the features learned from training well-design networks like ResNet and Inception using ImageNet dataset are generally transferable. This assumption is confirmed by the work ``What makes ImageNet good for transfer learning?'' from Huh et al. \cite{efros}. ImageNet is confirmed to be a good dataset to produce general convolution feature extractors. 

\subsection{Key Assumption 3: Convolution Channels Can be Ranked and Pruned}
In particular, we are assuming that not all the convolution channels are useful. Works on model compression and network pruning provide for us a solid experimental data for this assumption. 

Most of the work in convolution channel pruning involves a heuristic defined channel ranking methods, for examples: ranking channels by their $ell_1$ norms \cite{prune-l1-norm}, ranking channels by combinatorially determine their effect on validation score \cite{prune-combinatorial}, and the state-of-the-art ranking methods is first-order Taylor expansion with respect to the channel to closely match the cost function of original network \cite{prune-taylor}. 

Our approach is different. Instead of a human defined heuristic about image norm or Taylor expansion, we want the network to figure out how to rank filters by itself. We only provide the network with the an objective of classification score and a set of resource (convolution filters). 

\subsection{Key Assumption 4: Attention Should Be Regularized}
If we just assign the weight of $1$ to each convolution filters, the network will move very slowly or even remain stagnant. An important lesson we take from Kim et al. about visual attention is the importance of regularized attention \cite{canny-attn}. Although the work is about applying visual attention to self-driving cars, it provides important intuitions about regularizing visual attention and forcing it to look for new objects by tuning $\lambda$ as the strength of regularization. 

Our key take-away is just that adding attention in visual model is not enough. Especially in our model, where the attention is about \emph{the hierarchical relationship and strength of the convolution channels}, just adding a scalar weight of $1$ and freeze the convolution filter weights are not good enough. We need to add some prior about the attention weights. We further experiment with and discuss the different forms of regularization in section \ref{sec:attn-loss}.

\section{Experimental Setup}\label{sec:setup}
\subsection{Data Source and Pre-Processing}
\begin{figure}
    \centering
    \includegraphics[width=0.4\linewidth]{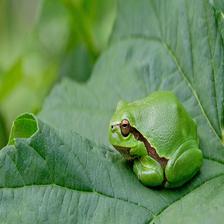}
    \includegraphics[width=0.4\linewidth]{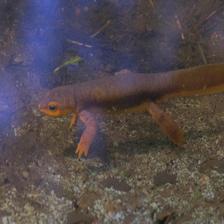}
    \caption{Example of iNaturalist Dataset}
    \label{fig:data-eg}
\end{figure}

We chose a subset of the INaturalist 2018 Dataset, in which there are 11,156 images of 144 species of amphibians. This dataset resembles a specific fine-grained classification dataset typically encountered in real life. It has quite a significant class imbalance for some classes, and significantly different data distribution from that of ImageNet. The amphibians share many common features and can be extremely similar among different species. Since they were usually photoed in their natural habitats, it is indeed a very hard dataset, yet general and diverse enough to represent a specific fine-grained classification task (example: Figure \ref{fig:data-eg}).

The images originally come in different sizes and shapes, and all of them are reshaped without cropping to 224x224 for ResNet models and 299x299 for Inception-V3. They are then normalized according to standard ImageNet practice, with $\mu =\begin{bmatrix}
           0.485 & 
           0.456 &
           0.406
    \end{bmatrix}$ and $
\sigma = \begin{bmatrix}
           0.229 &
           0.224 &
           0.225
    \end{bmatrix}$

Data augmentation was used to solved the problem of class imbalance and over-fitting in some experiments, which will be further discussed in section \ref{sec:data-augmentation}.

Data is split 7-1.5-1.5 for training, validation, and testing respectively. 

\subsection{Tools}
We used PyTorch\cite{pytorch} as our Machine Learning library for this project. We chose PyTorch over TensorFlow or other popular library because of its flexibility. PyTorch uses dynamic computational graph rather than a static one. 

In addition, PyTorch provides a simple way to customize pre-trained models to facilitate our experiment. For example, to freeze/unfreeze a specific variable inside the model, one can simply set .requires\_grad to desired boolean value; adding Attention weight to a specific channel also only requires minor changes to the existing pre-trained models.

All the pre-trained models used in our experiments (ResNet-\{34,50,101\}, Inception-V3) come from PyTorch Model Zoo \cite{pytorch-modelzoo}.

Models were mostly trained on EC2 instances using Tesla K80 GPUs, and some were trained on Pascal Titan X.

\subsection{Baseline Model}
We attempted different popular models pre-trained on ImageNet. Most experiments were done using ResNet 50, but ResNet 34,  ResNet 101 and Inception-V3 were also used to investigate compatibility of our method with different pre-trained model structures, and model depths. More on this in section \ref{sec:disc}. We showed that our attention module and training strategies are generally applicable to models with different architectures and depths. 

\subsection{Evolution of our model}
In order to solve the problem of filter redundancy in transfer learning on specific datasets, we first explored the idea of pruning, which ranks the convolutional filters based on some heuristic algorithms, and removes ones with less activation. Such a heuristic algorithm was proposed in \cite{prune-taylor}, which uses Taylor expansion to approximate the loss function and compares the loss when a certain filter is masked out to zero. This gives a heuristic estimation of the impact of this filter. However, we want to  propose an end-to-end method that uses attention weights as a measure of usefulness of a convolutional filter. The soft attention amplifies useful signals and suppresses redundant ones, which achieves a similar effect to pruning.

\section{Final Model}\label{sec:final-model}
\begin{center}
    \includegraphics[width=\linewidth]{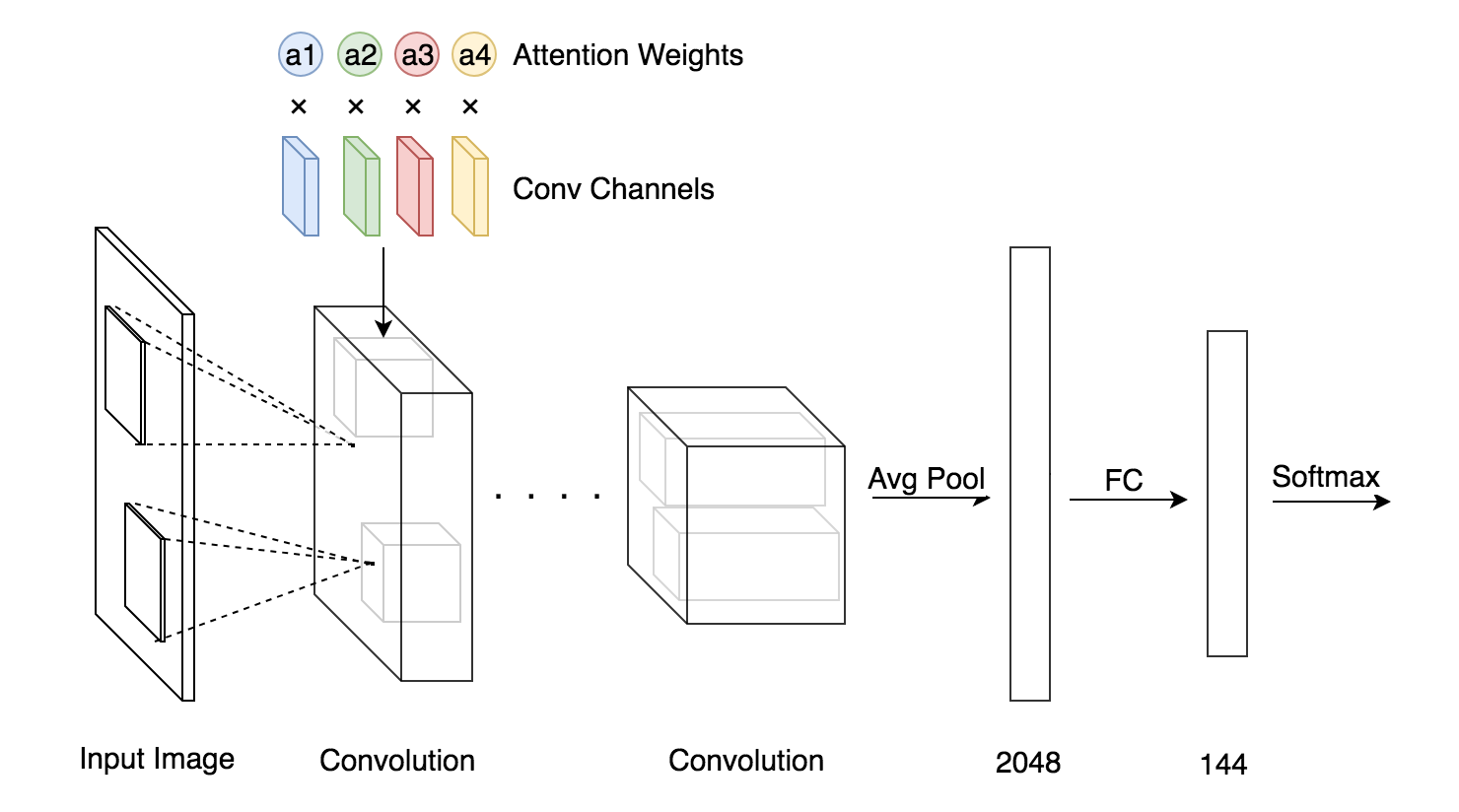}
\textit{Model Architecture}
\end{center}

The final model has attention modules attached at the output of each of the convolutional layers of the pre-trained models. This is done by multiplying each pre-trained convolution filter by a trainable scalar weight initialized at $1$. 

One to two epochs of the FC layers were trained to adjust the classifier to the fine-grained dataset, so that errors can be better back propagated to the attention modules. Next we fixed the entire network, and only trained the attention modules to identify the feature extractors in the pre-trained network that are useful for the specific fine-grained task. Training of the batchnorm layers was interspersed within training of the attention modules, in order to fight overfitting and adjust for variance shift. 

\subsection{Pseudocode}
The core of our model boils down to the pseodo-code shown below: 

\begin{verbatim}
# We initialize attention weights 
# as follows and fill it with 1
attn_weights = Parameter(FloatTensor(
out_channels, in_channels, 1,1))
...
# And we modify the "forward" function 
# in 2D convolution module
attn_paid = conv_weight*attn_weights
return F.conv2d(input, attn_paid, ...)
\end{verbatim}

\section{Experiment Result and Discussion}\label{sec:disc}
In this section, we describe a set of experiments we ran concerning many facets of fine-grained transfer learning and the specifics of our method. For each experiment, we will describe our setup, our result, and our findings. 

The investigation section contains experiments that were successfully ran and worked as expected. The visualization contains result from our experiment with interpretability of attention models. The surprise section consists of experiments that either gave us surprising results or just did not work out.

\subsection{Investigation}

\subsubsection{What is the best attention loss/regularization?}\label{sec:attn-loss}
We wanted to identify useful filters from redundant ones, and thus applied three different loss for attention weights. Initially, most attention weights clustered near 1, and were reluctant to differentiate filters. L1 or L2 norms of the attention weights were added to the loss function as regularization. L1 norm resulted in a sparser and more polarized distribution of the attention weights, but L2 converged slightly faster. Both achieved similar performance in terms of accuracy. 

\begin{equation}
    \ell_{1}(a_j) = \|a_j\|_{1}
\end{equation}

\begin{equation}
    \ell_{2}(a_j) = \|a_j\|_{2} 
\end{equation}

 In order to further help the attention weights diverge, we proposed a modified L2 loss that penalizes the distance of each attention weight to 1, which also serves as regularization. This penalty scheme was most commonly used in our model.

\begin{equation}
    \ell_{3}(a_j) = -\|a_j-1\|_{1}^{2} 
\end{equation}

The loss function is thus:
\begin{equation}
    \mathcal{L} = - \sum_{i=1}^{n} y^{(i)} \log(p)^{(i)} + \lambda \cdot \sum_{j=1}^{F}\ell(a_{j})
\end{equation}
where F is the number of filters in the convolutional layers, $a_{j}$ is the attention weight vector for filter j, and $\lambda$ is a hyper-parameter.

\begin{table}[]
    \centering
    \begin{tabular}{ |p{1.5cm}||p{1.5cm}|p{1.5cm}|p{2cm}|  }

 \hline
 Loss& Top 1 Val&Top 3 Val& Best Epoch\\
 \hline
L1   & 38.6\%    &60\%&   8\\
 \hline
 L2   & 42.7\%    &63.4\%&   10\\
 \hline
\end{tabular}
    \caption{Result from different loss functions, using same FFAAABAAABAA training scheme. F: fully connected layers; A: attention layers; E: all layers other than attention layers; B: batch norm layers}
    \label{tab:table_loss}
\end{table}

\begin{figure}
    \centering
    \includegraphics[width=\linewidth]{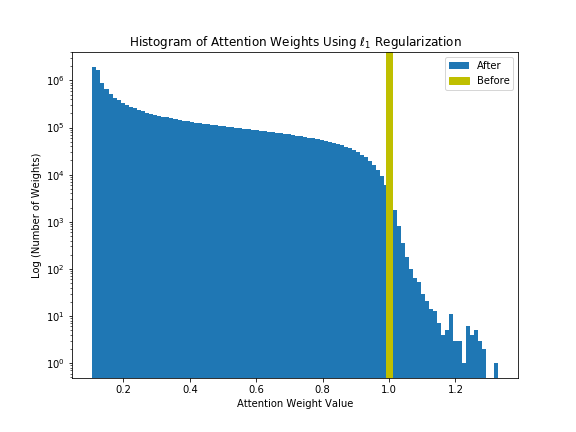}
    \includegraphics[width=\linewidth]{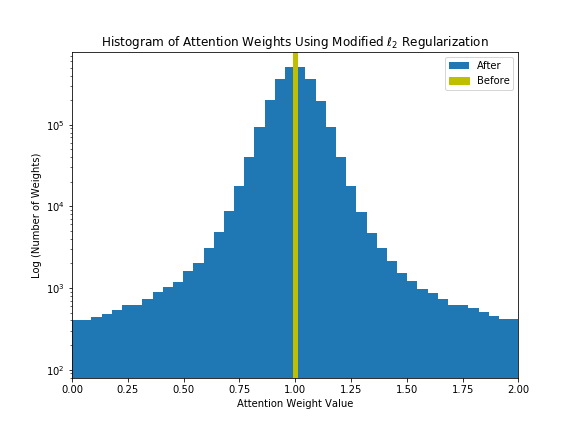}
    \caption{Different Attention Regularization, the wider the better}
    \label{fig:hist}
\end{figure}

As figure \ref{fig:hist} shows, both regularization scheme/prior lead to expected result distribution. The $\ell_1$ regularization put a sparsity inducing prior on attention weights; the result shows many weights are reduced close to zero. Surprisingly, there are few attention weights that are still amplified (value $>1$). We expect these weights help identify the subset of convolution channels that are useful for fine-grained classification with our dataset. The histogram for $\ell_2$ attention loss show similar but smoother result. The distribution of weights gradually diverge from $1$ and shift towards either $0$ or $2$ according to their significance in this particular fine-grained classification task. 

Our result is as shown in Table \ref{tab:table_loss}.






\subsubsection{Does data augmentation help?}\label{sec:data-augmentation}
Data augmentation was done with a weighted sampler that samples from each class with probability inverse to the number of training images in that class. This solves the problem of class imbalance and also serves as a regularizer to combat overfitting.

\begin{table}[]
    \centering
\begin{tabular}{ |p{1.5cm}||p{1cm}|p{1cm}|p{1cm}|p{1.5cm}|  }

 \hline
 Training Scheme& Top 1 Val&Top 3 Val& Epoch& Data Aug\\
 \hline
 FC*10   & 28.9\%    &46.6\%&   10& No\\
  \hline
 FC*10   & 28.5\%    &48.4\%&   10& Yes\\
 \hline
 FC+Att   & 42.7\%    &63.4\%&12&   No\\
  \hline
 FC+Att   & 50.5\%    &\textbf{70.0\%}&12&   Yes\\
 \hline
\end{tabular}
    \caption{We used some fixed training schemes with good results for controlled experiment.* FC+Att: FFAAABAAABAA}
    \label{tab:data_aug}
\end{table}

Our result is as shown in Table \ref{tab:data_aug}. Using data augmentation does improve the performance and it works particularly well with attention modules.

\subsubsection{How Does Different Model Architecture Affect the Score?}

Our result is as shown in Table \ref{tab:table_models}. We expect the validation accuracy to be about the same across \underline{different depth} of the network as well as \underline{different model architecture}. 

As the depth of ResNet increase, we see increase in validation accuracy; this result is justified by the fact that there are much more convolution channel as the network goes deeper as well as more interesting intermediate features ready to be combined. 

If we switch the base architecture to a different comparable ImageNet model, Inception V3, we see the performance is not degraded. This show the compatibility of our method with any ImageNet models; as long as it contains convolution.

\begin{table}[]
    \centering
\begin{tabular}{ |p{2cm}||p{1.5cm}|p{1.5cm}|p{1.5cm}|  }

 \hline
 Model& Top 1 Val&Top 3 Val& Best Epoch\\
 \hline
ResNet34   & 39\%    &60\%&   9\\
 \hline
RestNet50   & 43\%    &63\%&   10\\
 \hline
RestNet101   & 47\%    &66\%&   10\\
 \hline
Inception V3   & 41\%    &62.7\%&   5\\
 \hline
\end{tabular}
    \caption{Result from different model architecture, using same FFAAABAAABAA training scheme.}
    \label{tab:table_models}
\end{table}

\subsubsection{Is The Accuracy Good Enough (Compare with 100 hours models)?}
There is a baseline model \cite{pre-trained-inat-github}  trained from scratch using Inception-V3 on a larger and a more complete dataset of animal and plant species, of which our amphibian dataset is a subset. It took 100 epochs for the  model to achieve a final top 3 accuracy of $77\%$. 

In comparison, our model, though trained on a more specific dataset, were able to achieve the highest $70\%$ top 3 validation accuracy in just under 15 epochs, which took around \emph{13 minutes} on a Pascal Titan X GPU to train. 

Thus, the attention model is more suitable for classification tasks where high accuracy is not the absolute priority but require fast iteration of training and decent results. This makes it possible to train a large number of personalized models with personal data.

\subsection{Interpretability and Visualization}
In this section, we present two visualizations that showcase the effectiveness of our method for picking up good signals. 

    \begin{figure}[h]
    \centering
    \includegraphics[width=\linewidth]{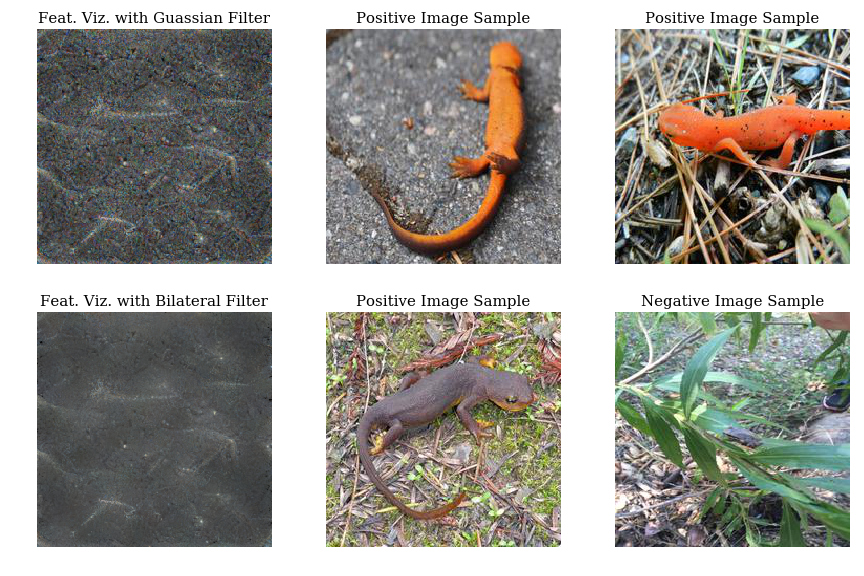}
    \caption{Direct Feature Amplified by Attention}
    \label{fig:direct-feature}
    \end{figure}

    \begin{figure}[h]
    \centering
    \includegraphics[width=\linewidth]{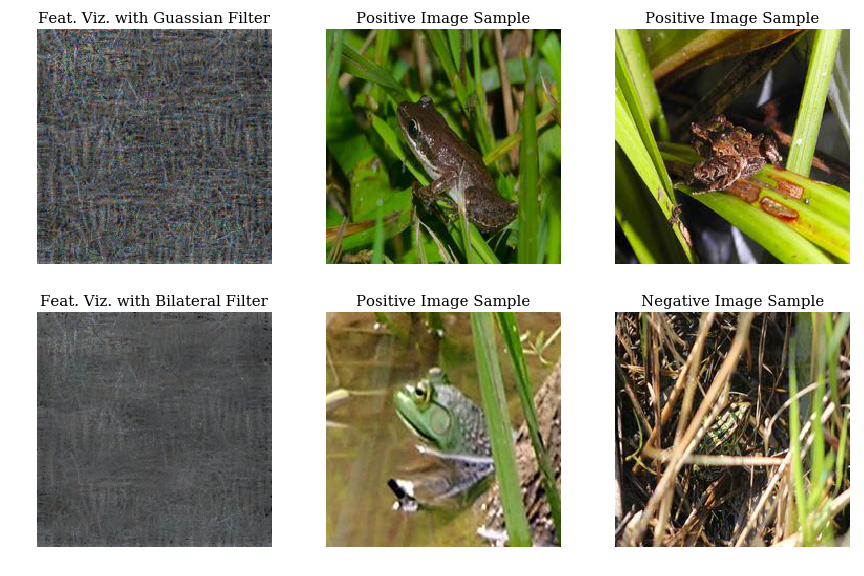}
    \caption{Auxiliary Feature Amplified by Attention}
    \label{fig:aux-feature}
    \end{figure}

We visualize two channels (Figure \ref{fig:direct-feature}, Figure \ref{fig:aux-feature}) in the third to last convolution layer of ResNet 50 (\texttt{res5c\_branch2a}). The channels are the top-2 channels ranked by the ``attention'' weights on such channels after training attention weights based on ResNet50 for few iterations using $\ell_2$ penalty to regularize the attention weights. Convolution features are visualized by optimizing random noise image to maximize the activation of a convolution channel (activation measured using mean of the image, applied Gaussian blur or Bilateral Filter, described in \cite{olah2017feature}). The images are sampled from top-10 images that maximally activate the channels ranked by $\ell_1$ norm of the feature map, the metric as inspired by convolution filter pruning method in \cite{prune-l1}. 

\emph{What does these visualizations shows?} 

Figure \ref{fig:direct-feature} shows a high level feature of spines of animals. The attention method is able to pickup the specific feature that significantly helps classification for this fine-tuned dataset. We also included a negative image sample. The leaf structure strongly resembles the spine structure therefore also activates this filters. 

Figure \ref{fig:aux-feature} shows another high level feature. However, this feature is an auxiliary feature that helps identify the surrounding environment of different types of frogs. The vertical leaf structure was picked up as a strong signal that can assists the network in identifying species.

\subsection{Surprise}

\subsubsection{Attention shape}

To answer this question, we ran two sets of controlled experiments. For the "Out Only" method, we assigned the same attention weight to each input channel in the same 3D convolutional filter. For the "In $\times$ Out" method, we assigned a different weight to each input channel.

As a concrete example, the first convolution layer of the ResNet-Family model takes as input a depth 3 RGB representation of the image and outputs a depth 64 feature map, for method 1 "Out Only" there are 64 attention weights in this layer while method 2 "In $\times$ Out" have 3 $\times$ 64 attention weights.

\begin{table}[]
    \centering

        \begin{tabular}{ |p{1.5cm}||p{1.5cm}|p{1.5cm}|p{2cm}|  }

 \hline
 Attention Channel& Top 1 Val&Top 3 Val& Best Epoch\\
 \hline
Out Only   & 41.2\%    &60.3\%&   \textbf{5}\\
 \hline
In $\times$ Out   & 42.7\%    &63.4\%&   10\\
 \hline
\end{tabular}

    \caption{Result from different attention shape, using same FFAAABAAABAA training scheme.}
    \label{tab:shape}
\end{table}

Comparing the number of parameters in either model (in Table \ref{tab:shape}), we initially expect the second method to outperform the first one by a large advantage. However, our experiment result only reveals a relatively small boost of performance. This is surprising because we thought a more sophisticated attention scheme will result in a drastically higher performance. 

One possible explanation for this surprising result is the hierarchical connection between a single $H \times W$ convolution filter and it's filter group, where all filters in the group collectively result in just one single feature map:
\begin{equation*}
        \text{out}(N_i, C_{out_j}) = \sum_{k = 0}^{C_{in} - 1} \text{weight}(C_{out_j}, k) \star \text{input}(N_i, k)
\end{equation*}
where $\star$ is the convolution operator, $C_{out_j}$ is the $j$th feature map. A single feature map is produced by adding the convoluted result from each filter in the group via $\sum_{k = 0}^{C_{in} - 1}$. \emph{We hypothesize there is a strong connection between different filters in the same group}, making attention weight less effective. 

Despite of their similar performance, the "Out Only" method converges to the best model much faster than the "In $\times$ Out" method (5 versus 10 epochs). 

Our result is as shown in Table \ref{tab:shape}.

\subsubsection{Do Traditional Transfer Learning Methods Still Work?}

We also explored different levels of 'cut', which are widely used in traditional transfer learning methods. We define the level of 'cut' as the number of layers that were not frozen when training. We trained different blocks of the pre-trained network for different number of epochs, including (1) Fully Connected Layers, (2) Block 4 of ResNet, and (3) the entire network end to end.

Method (1) gives us a good linear transformation from fixed high level feature (after average pooling) to the probability of classes. However it is prone to over-fit within few epochs. The best top 3 validation score we got is 46.6\%.

Method (2) unfreezes the last large block of ResNet and attempts to re-train high level features. We found the accuracy to be slightly higher: 56.6\%

Surprisingly, method (3) uses the entire pre-trained network as initialization. The final best validation score is 65.05\%, which is better than attention methods. However, this can be justified by the fact that all elements in every convolution filter are subject to optimization. Amplifying a convolution activation by a scalar weight has less complexity than fine-tuning every filter.

\section{Lesson Learned}\label{sec:lesson-learned}

We investigated the effectiveness of channel wise attention in the context of transfer learning for fine-grained specific classification tasks. Compared to just re-training the fully connected layers, the attention modules are surprisingly effective in quickly identifying useful in and out convolution channels, and it raises the performance of the model to nearly $70\%$ top 3 validation accuracy in under 12 epochs. An inception-V3 model trained with 100 epochs achieved $77\%$ top 3 accuracy. Our attention modules indeed provides a fast and relatively high performance model for transfer learning. 

We did different investigations to find out the best training strategy, attention penalty, level of cut, as well as the effect of data augmentation. We used different training strategies and used different combinations of FC, Attention, and BN layers. We discovered that FC layers quickly adjusted the classifier to the fine-grained data distribution but overfits after a few epochs, whereas attention layers effectively boosts the accuracy in a few epochs after the FC layers are trained. After the attention layers, however, the output feature maps are no longer well-behaved. When training batchnorm layers that are interspersed in between attention layers, accuracy drops slightly due to its regularization effect, but they help attention modules to achieve better results. 

Furthermore, the attention weights would cluster around 1 when only using cross entropy loss, but the addition of attention penalty encourages them to diverge and form different distributions, which yields roughly equally good performance. We also found the effectiveness of attention in picking interpretable image features that can help explain the network's decision. 

Our project provides an intuitive approach for fine-grained transfer learning. We still have a long way to go in exploiting pre-trained networks and adapting them for fine-grained or even personalized computer vision models.

\section{Future Work}\label{sec:future-work}
There are multiple lines of future work possible beyond our current stage. \begin{enumerate}
    \item A natural continuation of the project is to prune convolution channels that have attention weights smaller than a certain threshold to reduce the size of the final model as well as to improve accuracy. Attention weights provide us with a good ranking of features available for pruning. 
    \item Another direction is to further exploit the benefit of interpretability that attention scheme brings to the table. The attention weights are highly interpretable. 
    \item In our current approach, we have to experiment with different architectures and training schemes to find a satisfiable configuration. A natural step beyond this project will be to use reinforcement learning, especially meta learning strategies to let the model to learn the best way of integrating attention modules given a set of pre-trianed filters with the objective to classify fine-grained data correctly.
\end{enumerate} 

\section{Team Contributions}\label{sec:contribution}
Work was evenly split among the group. Simon came up the initial idea, the group refined the idea and designed experiment setups together.
Each of us experimented with multiple controlled experiments so that we can efficiently allocate GPU usage. We all worked on the poster, final report and code. \footnote{https://github.com/simon-mo/attn-attn}



\end{document}